\documentclass[twoside,11pt]{article}

\usepackage[accepted, specialissue]{melba}

\usepackage{amsmath,amsfonts}
\usepackage{multirow}

\melbaid{2023:012}  % This is provided upon by the publishing editor
\volume{2}
\firstpageno{338}  % Communicated by the publishing editor
\doi{https://doi.org/10.59275/j.melba.2023-553a}

\melbaauthors{Rasoulian et al.}  % Note: this one is also used to set the pdf 'authors' metadata
\melbayear{2023}  % The publication year
\datesubmitted{04/2023}  % Date submitted to MELBA: mm/yyyy
\datepublished{08/2023}  % Today's date: mm/yyyy

\melbaspecialissue{Machine Learning in Clinical Neuroimaging (MLCN) 2022}
\melbaspecialissueeditors{Seyed Mostafa Kia, Mohamad Habes}

\ShortHeadings{Weakly Supervised ICH Segmentation}{Rasoulian et al.}

\title{Weakly Supervised Intracranial Hemorrhage Segmentation using Head-Wise Gradient-Infused Self-Attention Maps from a Swin Transformer in Categorical Learning}

\author{\name Amirhossein Rasoulian \orcid{0009-0004-9077-521X} \email ah.rasoulian@gmail.com \\  % start right after \author{, or there will be an extra space
	\addr Department of Computer Science and Software Engineering, Concordia University, Montreal, QC, Canada
	\AND
	\name Soorena Salari \orcid{0000-0002-2587-0323} \email soorena.salari374@gmail.com \\
	\addr Department of Computer Science and Software Engineering, Concordia University, Montreal, QC, Canada
     \AND
	\name Yiming Xiao \orcid{0000-0002-0962-3525} \email yiming.xiao@concordia.ca \\
	\addr Department of Computer Science and Software Engineering, Concordia University, Montreal, QC, Canada
}

\begin{document}

% top matter
\maketitle

% abstract
\begin{abstract}%   <- trailing '%' for backward compatibility of .sty file
Intracranial hemorrhage (ICH) is a life-threatening medical emergency that requires timely and accurate diagnosis for effective treatment and improved patient survival rates. While deep learning techniques have emerged as the leading approach for medical image analysis and processing, the most commonly employed supervised learning often requires large, high-quality annotated datasets that can be costly to obtain, particularly for pixel/voxel-wise image segmentation. To address this challenge and facilitate ICH treatment decisions, we introduce a novel weakly supervised method for ICH segmentation, utilizing a Swin transformer trained on an ICH classification task with categorical labels. Our approach leverages a hierarchical combination of head-wise gradient-infused self-attention maps to generate accurate image segmentation. Additionally, we conducted an exploratory study on different learning strategies and showed that binary ICH classification has a more positive impact on self-attention maps compared to full ICH subtyping. With a mean Dice score of 0.44, our technique achieved similar ICH segmentation performance as the popular U-Net and Swin-UNETR models with full supervision and outperformed a similar weakly supervised approach using GradCAM, demonstrating the excellent potential of the proposed framework in challenging medical image segmentation tasks.
	Our code is available at ~\url{https://github.com/HealthX-Lab/HGI-SAM}.
\end{abstract}

% keywords

% Introduction (or first section)
\section{Introduction}

Intracranial Hemorrhage (ICH) is a potentially fatal cerebrovascular disorder that is responsible for 10-15\% of all stroke cases and can be caused by various factors, such as head trauma, high blood pressure, and blood clots \citep{rajashekar2021intracerebral,apostolaki2021prognosis}. The outcome of ICH depends on the volume of bleeding, which can enlarge rapidly within the first few hours \citep{qureshi2011antihypertensive}, leading to a high risk of secondary brain injury or even death if it is not treated promptly. In general, ICH can be classified into five subtypes based on its location in the brain, including Intraventricular (IVH), Intraparenchymal (IPH), Subarachnoid (SAH), Epidural (EDH), and Subdural (SDH). Note that one patient may have more than one hemorrhage subtype. Each ICH subtype should receive customized treatment approaches, and surgery is only considered if the location of the hemorrhage is advantageous. Upon admission at the hospital, early detection and accurate quantification of ICH are critical in selecting appropriate medical interventions and reducing patient mortality. Thus, efficient and automated systems to asses ICH are highly valuable. Compared to other medical imaging modalities, such as MRI, computerized tomography (CT) is often used in the clinic to assess ICH due to its fast imaging time and good accessibility. However, in addition to the morphological and spatial variabilities, the subtle contrast of ICH within often noisy clinical CT scans can pose challenges in its detection and quantification.

Recent progresses in deep learning (DL) techniques, especially convolutional neural networks (CNNs), have led to the development of efficient and accurate solutions for computer-assisted diagnosis and treatment decisions. For the care of intracranial hemorrhage, several automatic CNN-based DL algorithms have been devised for the detection, subtyping, and volumetric segmentation of intracranial hemorrhage based on clinical scans \citep{hssayeni2020intracranial,alis2022joint,Salehinejad2021ARD}. 
To overcome the limitations of CNNs in encoding long-range spatial information due to limited field of view, which may impact the accuracy of ICH detection and subtyping, particularly in cases where the spatial location of hemorrhage is crucial for diagnosis, the Vision Transformer (ViT) \citep{dosovitskiy2020image} has emerged as a promising solution. The ViT utilizes multi-head attention mechanisms to capture contextual relationships among spatially distributed image patches and has attracted great interest for vision tasks, including medical imaging applications \citep{dai2021transmed,dalmaz2022resvit}. However, by removing convolutions, the ViT possesses low locality inductive biases, such as translation invariant features. To address this, a recent variant called the Swin transformer \citep{liu2021swin} was introduced as an efficient hierarchical transformer, addressing the need for both long-range spatial encoding and local feature representation. It achieves the goal by gradually reducing the number of tokens by merging image patches and computing attention in non-overlapping local windows to mitigate the drawback of the ViT.

Training CNNs and Transformer-based models require a significant amount of data, but annotating medical images is a laborious and time-consuming process, particularly for segmentation tasks. Among various strategies, including semi-supervised learning, weakly supervised methods \citep{Zhou2017Review} offer alternative solutions to address such challenges by deriving fine-grained image segmentation from coarse and more accessible image annotations, such as bounding boxes, scribbles, and categorical labels. Among these typical choices, as categorical labels require the least time and effort, obtaining pixel-wise segmentation from them is highly attractive. This is especially true for our target application, where image classification is also needed, but such approaches have rarely been attempted. In this study, we intend to propose and validate a novel weakly supervised ICH segmentation technique by taking advantage of the Swin transformer.

In our previous work \citep{rasoulian2022weakly}, we employed a Swin transformer to perform CT-based detection and weakly supervised segmentation of ICH for the first time. More specifically, we obtained ICH segmentation by fusing hierarchical self-attention maps generated from a Swin transformer that was trained using categorical labels for ICH detection. Furthermore, comparing the proposed weakly supervised ICH segmentation framework for two Swin transformers based on (1) binary classification (presence of hemorrhage or not) and (2) multi-label classification (detailed ICH subtypes and with/without ICH), we found that binary classification helped better focus the network attention on the ICH regions. In this paper, we further extended our previous study \citep{rasoulian2022weakly} with three main contributions. \textbf{First}, inspired by the gradient-weighted class activation mapping (Grad-CAM) \citep{selvaraju2017grad}, we proposed a novel attention visualization technique, called HGI-SAM (Head-wise Gradient-infused Self-Attention Mapping), by performing head-wise weighing of self-attention obtained from the Swin transformer using the gradient of the target class. We further demonstrated the benefit of incorporating HGI-SAM in our weakly supervised ICH segmentation framework over the original proposal \citep{rasoulian2022weakly}. \textbf{Second}, by inspecting the characteristics of the gradient-weighted attention maps obtained from ICH detection, we proposed tailored post-processing methods to optimize the segmentation accuracy. \textbf{Lastly}, with the publicly available RSNA 2019 Brain CT hemorrhage \citep{flanders2020construction} and PhysioNet datasets \citep{hssayeni2020intracranial}, we conducted a comprehensive evaluation of the new method against our previous approaches, popular U-Net and Swin-UNETR models with full supervision, and a similar weakly supervised segmentation method leveraging the popular Grad-CAM technique, in the tasks of ICH segmentation and detection.

%%%%%%%%%%%%%%%%%%%%%%%%%%%%%%%%%%%%%%%%%%%%%%%%%%%%%%%%%%%%%%%%%%%%%%%%%%%
% Related works
%%%%%%%%%%%%%%%%%%%%%%%%%%%%%%%%%%%%%%%%%%%%%%%%%%%%%%%%%%%%%%%%%%%%%%%%%%%
% Make sure to put your work into context and include apporpriate citations.
% We do not have limits on citation counts.
\section{Related Works}
%%%%%%%%%%%%%%%%%%%%%%%%%%%%%%%%%%%%%%%%%
%A more thorough review of weakly supervised segmentation and attention/activation/saliency map mechanisms are neecded here + ICH segmentation/detection review
%%%%%%%%%%%%%%%%%%%%%%%%%%%%%%%%%%%%%%%%%%

There have been several variants of the Swin transformer model for medical image segmentation tasks. \cite{heidari2023hiformer} introduced a model with an encoder that combines feature maps from a CNN and a Swin transformer, to achieve accurate segmentation of skin lesions, multiple myeloma cells, and abdominal CT scans. \cite{cao2023swin} proposed a Swin-based U-Net-like model to segment abdominal CT images and cardiac MRI scans. \cite{hatamizadeh2022swin} proposed a hybrid Swin-encoder-CNN-decoder model to segment brain tumor MRI images. Finally, \cite{lin2022ds} introduced a dual Swin transformer model with different patch sizes to segment endoscopic images. Although all these methods showcase promising results to demonstrate the capability of the Swin transformer architecture, they all require full supervision.

To overcome the challenge of limited, well-annotated training data in developing deep learning techniques for medical image segmentation, a number of semi-supervised and weakly supervised algorithms have been proposed \citep{Wang2022Review,QURESHI2023316,syed2023weakly}. Semi-supervised strategies leverage a small number of images with refined labels, along with unlabeled or weakly labeled data. In this domain, \cite{yurt2022semi} used Generative Adversarial Networks (GANs) for MRI contrast translation with undersampled k-space data. \cite{chen2019multi} employed attention-based multi-task learning that simultaneously optimizes a supervised segmentation and an unsupervised reconstruction for brain tumor segmentation. Finally, \cite{zhou2019collaborative} incorporated collaborative learning for diabetic retinopathy grading and lesion segmentation. On the other hand, weakly supervised techniques rely entirely on coarse labels in the formats of bounding boxes \citep{Deepcut2017}, scribbles \citep{LIU2022108341}, points \citep{Roth2021}, or even categorical labels \citep{lin2018scannet}. As these coarse-level labels are more economical to acquire, weakly supervised segmentation techniques can further reduce the need for refined pixel/voxel-level annotations. With simple bounding boxes, \cite{Deepcut2017} proposed DeepCut, an approach that combined a CNN segmentation model with a densely-connected conditional random field (CRF) in an iterative training process to achieve pixel-level segmentation. Their method was tested on brain and lung segmentation for fetal MRI datasets. Following the approach, \cite{Kervadec2020BoundingBF} employed global constraints derived from box annotations, including tightness prior and global background emptiness, to achieve improved segmentation results over DeepCut \citep{Deepcut2017} on the PROMISE12 dataset \citep{litjens2014evaluation}. Previously, scribble and point annotations have been widely used in interactive segmentation. In weakly supervised segmentation, \cite{Roth2021} employed the random walker algorithm to generate coarse image-level labels from anatomical landmarks, which were used in combination with the point clouds to refine the segmentation results. More recently, \cite{LIU2022108341} proposed a weakly supervised COVID-19 infection segmentation method based on image scribbles and an uncertainty-aware mean teacher framework. 

To further alleviate the need for pixel/voxel-wise manual annotation, weakly supervised segmentation methods that solely rely on categorical labels are highly attractive. With the assumption that deep neural networks in image classification tasks should have a local focus on the target objects, this type of approach was made possible by the latest techniques that provide an intuitive visual explanation of the reasoning process for DL algorithms through saliency, class activation, and attention maps. In this domain, \cite{han2022weakly} proposed a weakly supervised segmentation model based on class residual attention for the lung adenocarcinoma and breast cancer datasets. \cite{chen2022c} developed a novel class activation mapping for weakly supervised segmentation for MRI datasets that achieves state-of-the-art accuracy, and similarly, \cite{viniavskyi2020weakly} utilized class activation maps (CAM) for Chest X-Ray segmentation. More recently, \cite{yu2022adaptive} further modified CAMs by scale feature adaptation and soft-erase modules to segment thyroid ultrasound images.
With the transformer model, \cite{li2023weakly} utilized a self-attention mechanism in multiple instances learning for weakly supervised segmentation of histopathology images while \cite{zhang2022transws} used CAM and a refinement segmentation decoder for the same task.

Almost all previous reports on automatic ICH detection and/or segmentation primarily relied on supervised learning strategies. \cite{hssayeni2020intracranial} recently conducted a comprehensive review of these techniques in both semi-automatic and automatic manners, and binary classification (ICH versus non-ICH) achieved an area-under-the-curve (AUC) of 0.846$\sim$0.975, while more fine-grained ICH subtyping achieved an AUC of 0.93$\sim$0.96. Deep learning-based approaches in ICH detection typically used fully convolutional networks (FCNs) \citep{Cho2018ImprovingSO} and recurrent neural networks (RNNs) \citep{Ye2019PreciseDO}, and their accuracy was generally higher for ICH versus non-ICH classification than for ICH subtyping. 
Following the trend in explainable artificial intelligence (XAI), attention mechanisms have been employed to both boost detection accuracy and visually illustrate classification results.
\cite{saab2019doubly} and \cite{Salehinejad2021ARD} utilized ResNet-like architectures for binary ICH detection with attention layers and Grad-CAM techniques, respectively, but they only visualized attention and class activation maps for qualitative assessment of their methods.
Furthermore, very limited attempts were also made to apply the attention/class activation in weakly supervised brain lesion and hemorrhage segmentation \citep{Wu2019MICCAI,nemcek2021weakly,liu2022mixed}. Specifically, \cite{Wu2019MICCAI} used refined 3D CAMs to segment stroke lesions from the Ischemic Stroke Lesion Segmentation (ISLES) dataset (multi-spectral MRI), and achieved a 0.3827 mean Dice score. \cite{liu2022mixed} used multi-scale CAMs and a Mixed-UNet model with two decoder branches on top of a VGG-based binary classification CNN. They trained the network based on a private MRI dataset and achieved a 0.56 mean Dice score for ICH segmentation on a small CT dataset. Likewise, \cite{nemcek2021weakly} found the location of ICH as bounding boxes in axial brain CT slices based on the regional extrema of attention maps acquired from a ResNet-like binary classification CNN. In their approach, a mean Dice of 0.58 was reached for the lesion bounding boxes. Unfortunately, to the best of our knowledge, aside from our earlier work \citep{rasoulian2022weakly}, self-attention, especially with a Swin transformer, has not yet been explored for weakly supervised ICH segmentation, and we intend to further improve our proposed framework to boost the performance.

\section{Methods}
\begin{figure}[t]
    \centering
    \includegraphics[width=\textwidth]{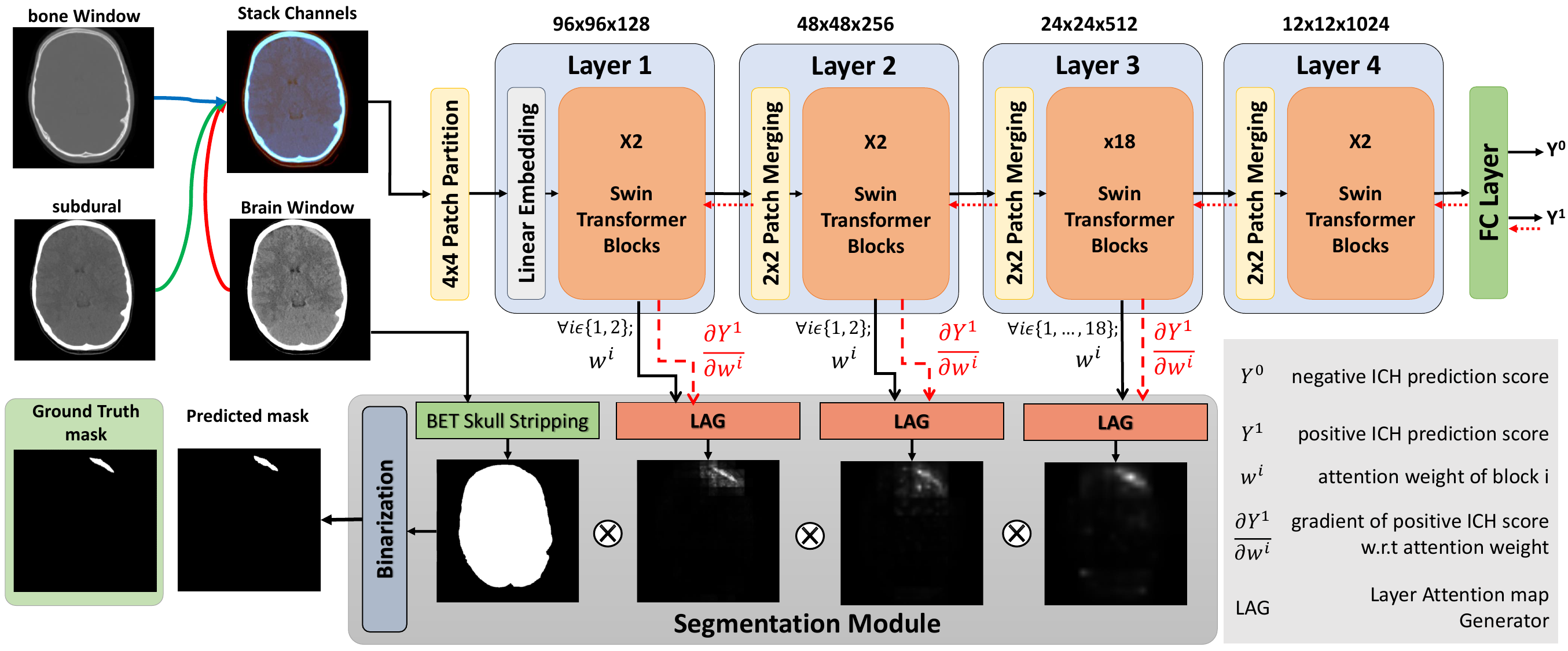}
    \caption{Overview of the proposed weakly supervised segmentation method using hierarchical fusion of gradient-weighted self-attention maps.} \label{model}
\end{figure}

An overview of our proposed weakly supervised technique for ICH segmentation is depicted in Fig.~\ref{model}, which comprises two major components. First, a Swin transformer was trained through an ICH detection task using categorical labels to classify input images into ICH vs. without ICH. Then, during test time, the segmentation module utilized hierarchical attention maps from the Swin transformer blocks along with their corresponding gradients to predict the hemorrhage segmentation map. Due to the high variability in slice thicknesses among the CT data, we decided to implement our algorithm based on 2D axial slices.
The details of the methodology are provided in the following sections.

\subsection{ICH detection with a Swin transformer}
In our proposed technique, we employed the Swin-Base transformer architecture, which divides an input image into $4\times4$ patches before passing their embedding through 4 layers/hierarchies to predict the existence of hemorrhages. Unlike the ViT, which computes the multi-head self-attention (MSA) between all image patches, in Swin-Base transformer, self-attention is derived within non-overlapping windows of $12\times12$ patches, which considerably reduces the computational cost. Here, for simplicity, we will refer to the Swin-Base transformer as ``Swin transformer" from this point on. Two main mechanisms help establish the associations between patches across different windows. First, the Patch-Merging module at the beginning of each Swin transformer layer combines and encodes every $2\times2$ neighboring patches into one. Second, every two consecutive transformer blocks apply window-based multi-head self-attention (W-MSA) and shifted window-based multi-head self-attention (SW-MSA) units to input tokens (see Fig.~\ref{layer_attention_map}a).
The self-attention per head within each window is computed as:
\begin{align}
    \label{eqn:attention}
    \begin{split}
        & \text{Attention}\left(Q_h, K_h, V_h\right) = w_h \times V_h , \\
        & w_h = \text{Softmax}\left(\frac{Q_h K_h^T}{\sqrt{d}} + B_h\right), \\
    \end{split}
\end{align}
where Q, K, and V denote query, key, and value vectors, respectively. $w$ is the window attention weight that we use to derive the attention map, $h$ denotes the head index of multi-head self-attention, $d$ is the dimension of the query or key, and B is the positional embedding matrix. Here, since the dimension of the window is $12 \times 12$, the dimension of $w$ is $144 \times 144$. Note that $w$ shows the relevance score of key tokens with query tokens. For more information on how the attention weight within each window is computed, we refer the readers to the original Swin transformer paper \citep{liu2021swin}.

In our earlier study \citep{rasoulian2022weakly}, we discovered that providing additional information (hemorrhage subtypes) to ``ICH vs. without ICH" classification during training can distract the network attention in the Swin transformer. As a result, for our new method with HGI-SAM, we decided to establish the backbone of our algorithm based on simple binary ICH detection. To benefit from the target class gradient, instead of using one output neuron to represent the classification outcome, we framed the final network with a two-class setup (i.e., positive and negative ICH detection). Further information on network training is detailed in Section 4.2.

\subsection{Hemorrhage Segmentation}
In our previous study \citep{rasoulian2022weakly}, we have qualitatively demonstrated the superior performance of self-attention maps than the class-activation maps obtained with Grad-CAM in visually explaining the ICH detection process in Swin transformers. Therefore, we continued to take advantage of self-attention maps, with a novel formulation to perform weakly supervised ICH segmentation. 

Previous attempts to visualize attention weights in the ViT involved inserting an extra classification token into the image patches and then extracting the attention weight of this token after multiplying the weights of all layers \citep{chefer2021transformer, dosovitskiy2020image}. However, this approach is not feasible for the Swin transformer due to its window division mechanisms for both regular and shifted windows. Additionally, multiplying different attention weights is challenging due to two reasons. First, at different layers/hierarchies, Patch-Merging results in a different feature map resolution and number of tokens. Second, every two successive Swin transformer blocks have attention weights corresponding to regular and shifted image patches that do not match. To address these challenges, we calculated the attention map at each block by averaging over all query tokens with additional operations of the window and shift reversal, and then the interpolated maps at different layers are multiplied.

\subsubsection{Layer Attention Map Generation}
\begin{figure}[t]
    \centering
    \includegraphics[width=\textwidth]{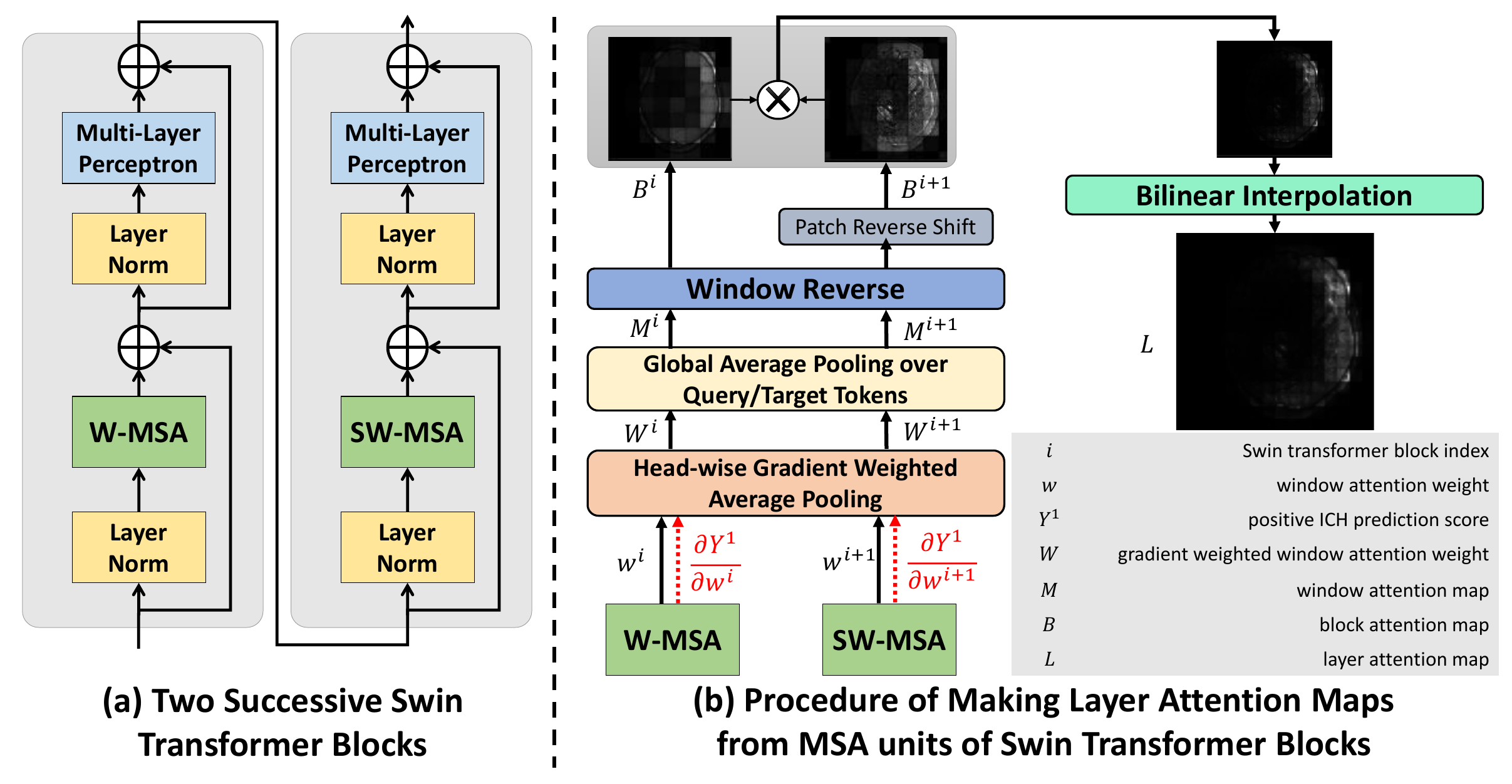}
    \caption{Demonstration of head-wise gradient-infused layer attention map generation in the proposed Swin transformer in categorical learning} \label{layer_attention_map}
\end{figure}
There has been recent research that leverages model classification scores for attention explainability. For instance, \cite{chefer2021transformer} utilize the Taylor Decomposition principle to assign and propagate a local relevance score through the layers of a ViT model. Similarly, \cite{sun2021getam} and \cite{barkan2021grad} employ attention gradient weighting on ViT and BERT models, respectively. However, these approaches primarily focused on the attention weight of the ``cls" token, and the latter two methods weighed each token's attention weight through element-wise multiplication. In contrast, our work places emphasis on weighing different heads in multi-head self-attention, and we performed the operation on the more complex Swin Transformer for the first time.

The use of multiple heads in the self-attention mechanism enhances the representational capacity and robustness of the transformer model, as each head can focus on different aspects of the input and learn a unique set of attention weights, thus capturing more complex relationships among the tokens. However, this critical fact was overlooked in most previous attention map generation methods \citep{gao2021ts}, including our own previous work \citep{rasoulian2022weakly}. In the existing literature, naive averaging is often applied to the attention weights of all heads to obtain an overall weight representation. However, as proved by \cite{voita2019analyzing}, some heads have more contribution to the output prediction. In this work, we weighed each head by the norm of its gradient regarding the classification score of positive ICH detection, which caused the attention weights of the heads that are more strongly associated with hemorrhage detection to have a heavier influence on the final attention weight representation. This is similar to Grad-CAM, where the target class gradient is used to weigh the associated activation map to enhance its specificity. In a Swin transformer, attention weights are computed within non-overlapping local windows while the W-MSA and SW-MSA units in two successive blocks establish cross-window connections. To encode the full attention information from local windows and cross-window connections, we multiply the attention maps from the original and shifted versions. Thus, we produce one map per every two consecutive blocks. As illustrated in Fig.~\ref{layer_attention_map}, the layer attention map is created as follows:

\begin{align}
    \label{eqn:layer_attention_map}
    \begin{split}
        & W^i = \frac{1}{H} \sum_{h=1}^{H}{\left\| \frac{\partial Y^1}{\partial {w^i}_h} \right\| \cdot {w^i}_h}, \\
        & M^i = \frac{1}{Q} \sum_{k=1}^{Q}{W^i_k}, \\
        & L = \text{BI} \left(
                                    \text{WR}\left(W^i\right)
                                    \otimes 
                                    \text{RS} \left(
                                                        \text{WR}(W^{i+1})
                                                \right)
                          \right)
    \end{split}
\end{align}
where $w^i$ is the Multi-head window Self-Attention (MSA) weight of block $i$, $H$ is the number of heads in the MSA unit, $M^i$ is the window attention map of block $i$ which is derived by averaging the window attention weight over its query tokens' dimension, and $L$ is the layer attention map. BI refers to bilinear interpolation, which is utilized to upsample the map to the image size. WR stands for Window Reverse operation, which involves concatenating maps of all windows to create a full image map. Also, RS denotes the reverse shift operation, which is used to reposition the shifted patches of the SW-MSA unit to their original locations in the image. It is good to mention that for Layer 3 in our Swin transformer, which consists of 18 blocks, the final layer map is obtained by averaging the results of 9 maps computed as above.

\begin{figure}[t]
    \centering
    \includegraphics[width=\textwidth]{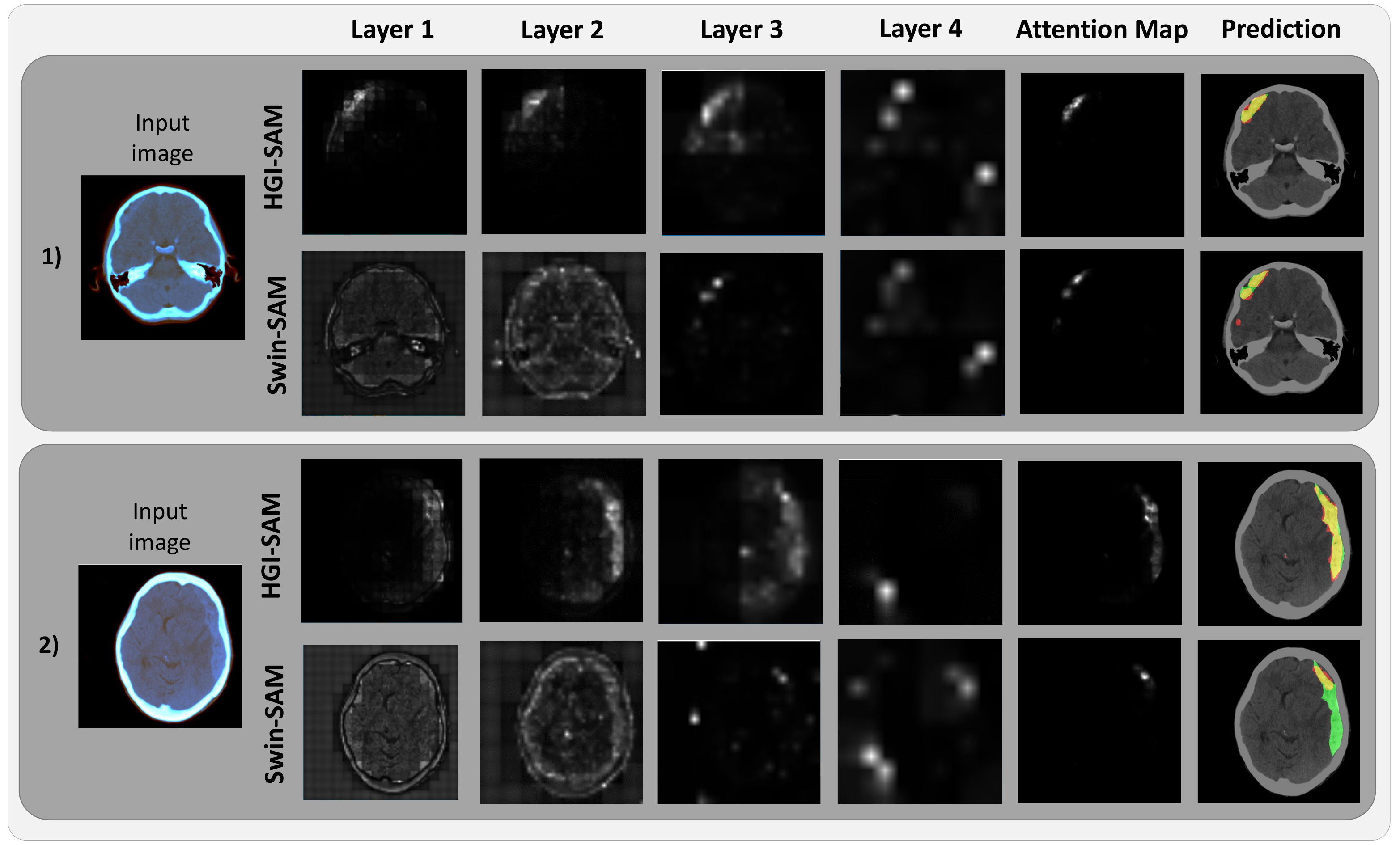}
    \caption{A comparison of the proposed head-wise gradient-infused self-attention mapping (HGI-SAM) and the original self-attention maps (Swin-SAM) from the Swin transformer model is shown for two axial CT slices. Along with the maps at different hierarchies, the fused attention maps and the derived binary ICH segmentation (in red) are also shown over the ground truths (in green). Note that the yellow color shows the overlapping area (true positive regions of segmentation results).} \label{attention_comparison}
\end{figure}

\subsubsection{Segmentation Module}
In the segmentation module, the final ICH segmentation was obtained by thresholding the pixel-wise multiplication result of the attention maps from different hierarchical layers in the Swin transformer. Note that the attention map generated by the last layer in the Swin transformer tends to be much more coarse due to the interpolation of a $12\times12$ pixel map to the image size, resulting in reduced resolution and potential loss of fine-grained details. Therefore, unlike our previous approach \citep{rasoulian2022weakly}, which used the attention of Layer 4 to compensate for its limited ability to capture relevant features in earlier layers, with the new technique using HGI-SAM, we used the attention maps from the first 3 layers to generate the final ICH segmentation \citep{Zhou2021DeepViTTD}. 
Furthermore, as demonstrated in Fig.~\ref{model}, we employed an additional post-processing step in our approach. This involved multiplying the final fused attention map with a brain binary mask, removing any irrelevant attention weights to ICH segmentation outside the brain region. The skull-stripping procedure was conducted following the recommended steps outlined by \cite{muschelli2019recommendations}. Lastly, the refined attention map was binarized using a simple thresholding method, which was demonstrated to be more robust than K-means or Otsu's method in our previous study \citep{rasoulian2022weakly}, resulting in a discrete segmentation mask. To determine the optimal threshold value, we conducted a grid search using the validation data. More specifically, to evaluate a fold in 5-fold cross-validation, we chose a best-performing threshold value from 0 to 1 with a step size of 0.01 that obtained the best segmentation on the remaining folds based on Dice scores.

\section{Experiments and Evaluation}
To investigate the performance of the proposed weakly-supervised ICH segmentation method using our new HGI-SAM technique, in addition to the approaches from our previous publication \citep{rasoulian2022weakly}, we also implemented three baseline models, including a fully supervised U-Net, a fully supervised Swin-UNETR, and a similar weakly supervised segmentation method based on binary ICH detection using class activation maps from Grad-CAM. To facilitate the discussion of these methods, we refer the weakly supervised segmentation techniques with Grad-CAM, self-attention maps in multi-label learning (ICH subtyping), self-attention maps in binary ICH detection, and head-wise gradient-infused self-attention maps in binary ICH detection as Swin-Grad-CAM, Swin-SAM Multi-label, Swin-SAM Binary, and Swin-HGI-SAM, respectively. All our networks were trained on a desktop computer with an Intel Core i9 CPU and an NVIDIA GeForce RTX 3090 GPU with 24 GB of memory. The following sections provide detailed information about the dataset, model training techniques for various segmentation methods, and evaluation metrics.

\subsection{Dataset}
 To train and evaluate our models, we used two public datasets, the RSNA ICH CT dataset \citep{flanders2020construction} and the PhysioNet CT dataset \citep{hssayeni2020intracranial}. The RSNA dataset contains 752,803 CT slices, with each slice annotated only with ICH subtypes, and the PhysionNet dataset has 2,814 CT slices (75 subjects) with both manual ICH segmentation and ICH subtypes. For all weakly supervised methods, the deep learning models were trained only using the RSNA dataset, which was randomly split into 90\% and 10\% for training and validation sets. We used the validation set to early stop training when its loss stops decreasing. Their testing was performed only using the PhysioNet dataset. To train and test the fully supervised U-Net and Swin-UNETR networks \citep{hatamizadeh2022swin}, subject-wise five-fold cross-validation was used on the PhysioNet dataset, where we ensure that no slices from the same subject exist across different folds. Finally, we incorporated the same data splitting to evaluate all techniques. We published our data splitting along with our code at ~\url{https://github.com/HealthX-Lab/HGI-SAM}.

 To prepare the data, for each CT slice, brain, subdural and bone windows created using the suggested parameters provided in the relevant data publications \citep{flanders2020construction,hssayeni2020intracranial} were stacked to create a three-channel image, downsampled to 384×384 pixels, and normalized using min-max scaling to the range of [0,1].

\subsection{Implementation details}
\subsubsection{ICH segmentation with Swin-SAM Multi-label and Swin-SAM Binary}
In our previous work \citep{rasoulian2022weakly}, two Swin transformers \citep{rw2019timm} were trained with categorical learning to provide self-attention maps for ICH segmentation, with one for binary ICH detection and the other for binary ICH detection and full subtyping. When training both models, we used the AdamW optimizer with an initial learning rate of 1e-5 and early stopping with a patience of 3 to avoid overfitting. To address the class imbalance (ICH vs. without ICH) issue in the dataset, we used the focal cross-entropy loss function. Finally, we employed data augmentation techniques, including random left-right flipping, image rotation, and Gaussian noise addition, to improve the capacity and robustness of the trained models. At test time using the PhysioNet data, the binary hemorrhage segmentation was obtained using the same post-processing step as described in Section 3.2.2. More specifically, a five-fold cross-validation approach was used to determine the optimal threshold to generate binary ICH segmentation masks from the fused attention maps. The fold-wise average of threshold values for Swin-SAM Multi-label and Swin-SAM Binary were 0.11 and 0.07, respectively. Additionally, the division of the five folds was made consistent with the training and testing of the supervised U-Net and Swin-UNETR models.

\subsubsection{ICH segmentation with HGI-SAM}
The Swin transformer for our new weakly supervised ICH segmentation using HGI-SAM was established based on that of the Swin-SAM Binary technique, following our previous insight regarding the benefit of binary classification on self-attention maps \citep{rasoulian2022weakly}. To allow the computation of class-specific gradients for HGI-SAM, instead of one neuron to represent binary ICH detection outcomes, the new model was equipped with two output neurons to represent the ICH positive class and ICH negative class. To take advantage of our existing work, the new model was fine-tuned based on the Swin transformer backbone of Swin-SAM Binary, using the AdamW optimizer with a learning rate of 1e-6 and early stopping. Here, data augmentation with random spatial transformations and Gaussian noise addition was used during training. Furthermore, with the cross-entropy loss function, during training, we adopted data sampling that drew training data samples with probabilities that were inversely proportional to their label frequencies to handle the class imbalance issue in the datasets. Upon completing the training, pixel-wise ICH masks were obtained in the same manner as described in the previous section to allow a consistent comparison for all techniques. The obtained fold-wise average threshold value for this model was 0.06.

\subsection{Baseline models}
\subsubsection{ICH segmentation with Grad-CAM}
As most existing weakly supervised segmentation techniques relied on Grad-CAM \citep{selvaraju2017grad}, we implemented a baseline technique of this category, where we employed the class activation map on the same Swin transformer that we trained with the binary ICH detection task for Swin-HGI-SAM. Following the suggestion by \cite{jacobgilpytorchcam}, we applied the Grad-CAM target layer to the output of the first norm layer in the final block of the Swin transformer. Similar to the proposed self-attention-based method, the activation map was first multiplied by the brain mask and then thresholded to achieve the final hemorrhage segmentation as described in Section 4.2. The obtained fold-wise average threshold value for this model was 0.80.

\subsubsection{Fully supervised U-Net}
The U-Net is one of the most popular DL models in medical imaging applications. Therefore, we implemented a fully supervised U-Net model with a lighter architecture than that in the PhysioNet ICH data paper \citep{hssayeni2020intracranial}, which has four hierarchies in the encoding and decoding paths, but less embedding dimension. Each hierarchy consists of two Convolutional layers with ReLU activation function, a Max-Pooling layer in the encoding branch, and Transposed Convolutional layer in the decoding branch. We used the AdamW optimizer with an initial learning rate of 1e-3, the same sampling and augmentation strategy as our weakly supervised models, and a loss function made of Dice coefficient and cross-entropy, in a five-fold cross-validation setup.

\subsubsection{Fully supervised Swin-UNETR}
Swin-UNETR \citep{hatamizadeh2022swin} is one the most popular Swin-based segmentation models that takes advantage of Swin transformer and CNN techniques at the same time. Specifically, it is a U-Net-like architecture, where the encoder is a Swin transformer, the decoder is a CNN, and skip connections pass through convolutional residual blocks. To mitigate overfitting considering the size of the PhysioNet dataset, we adopted a lighter version of its original model that has 4 layers/hierarchies, an initial embedding dimension of 12, and 2, 4, 8, and 16 heads in multi-head self-attention units of Layer 1 to 4. Here, we used the same training parameters and strategies as the U-Net model, which also offers the best outcome for this method, to train the network.

\subsection{Evaluation metrics}
For all the proposed and implemented methods, we evaluated their segmentation performance using Dice coefficient and Intersection over Union (IoU). In addition, to assess the performance of binary ICH detection, we also computed a range of metrics, including accuracy, area under the curve (AUC), precision, F1-score, recall, and specificity for all algorithms. Note that for Swin-SAM Multi-label, where the designated Swin transformer was trained for both binary ICH classification and subtyping, the performance was assessed only based on the binary detection results. For the U-Net and Swin-UNETR models, ICH detection was recorded as whether the network provided a hemorrhage segmentation for a given image since a similar approach was also used for assessing aneurysm detection in the ADAM MICCAI Challenge \citep{timmins2021comparing}. It is worth mentioning that to make the performance of these models in ICH detection more robust, we do not consider tiny foregrounds as positive ICH ($<$ 10 pixels). As the data division for five-fold cross-validations for different techniques was the same, we reported the ICH segmentation and detection accuracy for all folds. We also report the model's overall performance by considering the accuracy of all slices. Lastly, two-sided paired sample t-tests were performed to further confirm the performance of our newly proposed segmentation method based on HGI-SAM against the rest of the comparing group.

\section{Results}
\begin{figure}[ht]
    \centering
    \includegraphics[width=\textwidth]{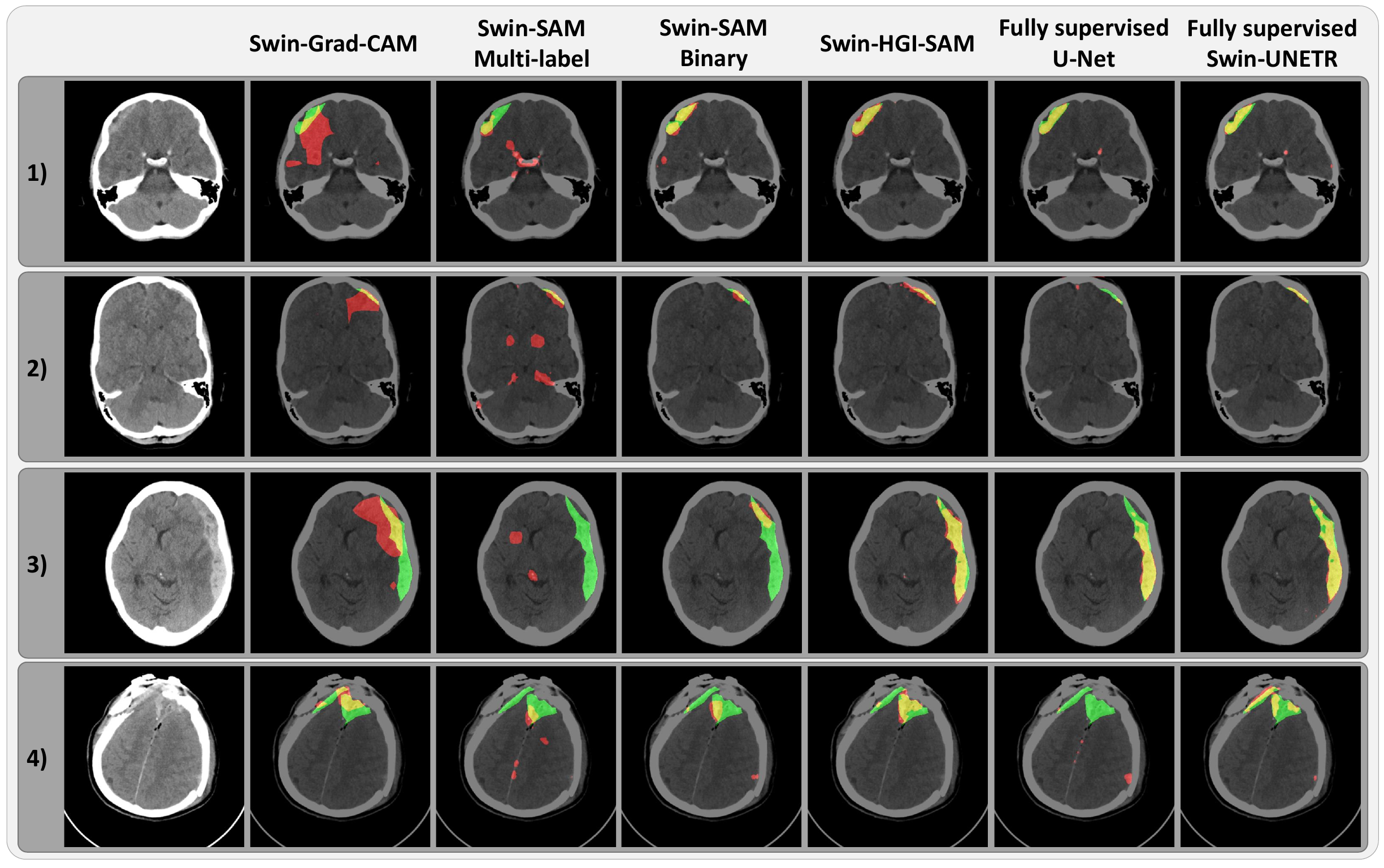}
    \caption{Qualitative comparison of segmentation performance for the proposed weakly supervised ICH segmentation methods (Swin-Grad-CAM, Swin-SAM Multi-label, Swin-SAM Binary, and Swin-HGI-SAM), fully supervised U-Net, and fully supervised Swin-UNETR for four different cases. Here, red=automatic segmentation, green=ground truths, and yellow=true positives.} \label{segmentation_comparison}
\end{figure}

To demonstrate the impact of gradient-weighing for self-attention maps and thus the final hemorrhage segmentation, we illustrate the layer-wise attention maps, along with the combined map and the binary segmentation in Fig.~\ref{attention_comparison} for the axial CT slices of two patients. From Fig.~\ref{attention_comparison}, it is evident that the proposed head-wise gradient-infused self-attention maps (HGI-SAM) provided more attention weights with higher specificity for the hemorrhage regions, especially at the first two layers with higher resolutions. This, in turn, provided final binary segmentations with a better agreement with the ground truths. To showcase the segmentation performance of the proposed method, the results of all mentioned techniques are shown for four different patients in Fig.~\ref{segmentation_comparison}. When comparing Swin-Grad-CAM and the self-attention-based results, we can see that while Swin-Grad-CAM could focus on the general region-of-interest correctly, it often provided much larger segmentations than needed. Between Swin-SAM Multi-label and Swin-SAM Binary, as we discovered in the previous study \citep{rasoulian2022weakly}, binary classification helped better focus the model attention in the hemorrhage region than the multi-label counterparts, thus offering more accurate segmentation. Finally, in contrast to the rest of the weakly supervised methods, Swin-HGI-SAM gave the most similar results to the fully supervised models, and notably, in Cases 2 and 4, the U-Net missed the small ICH that Swin-HGI-SAM and Swin-UNETR were able to identify.

Following the qualitative demonstration of the segmentation performance, the Dice coefficient and IoU metric for all methods are listed in Table~\ref{tab:seg_results} for all five folds from the experiments, with their overall slice-wise mean±SE. While the Swin-UNETR achieved a Dice of 0.455±0.019 and an IoU of 0.355±0.016 in a fully supervised setting, Swin-HGI-SAM was able to offer the second best results, with a 0.444±0.014 Dice. With Swin-Grad-CAM as the worst method, Swin-SAM multi-label and Swin-SAM binary performed worse than the newly proposed technique. In terms of statistical tests for segmentation metrics, Swin-HGI-SAM outperformed all weakly supervised methods ($p=10^{-37} < 0.05$ compared with Swin-Grad-CAM, $p=10^{-29} < 0.05$ compared with Swin-SAM Multi-label, $p=0.0029 < 0.05$ compared with Swin-SAM Binary) while producing similar segmentation accuracy as the fully supervised U-Net ($p=0.829 > 0.05$) and fully supervised SwinUNETR ($p=0.6184 > 0.05$)

Finally, in Table~\ref{tab:detection_results}, we listed the full assessment of ICH detection for Swin-SAM multi-label, Swin-SAM binary, Swin-HGI-SAM, U-Net, and Swin-UNETR. Despite the strong performance of fully supervised U-Net and Swin-UNETR models in ICH segmentation, their ICH detection accuracy falls short when compared to weakly supervised models trained with categorical labels. For all Swin transformer models, they offered similar ICH detection performance across all evaluation metrics. By comparing Table~\ref{tab:seg_results} and Table~\ref{tab:detection_results} across different data folds, we noticed that the detection results align with segmentation performance, especially for weakly supervised based models. This is expected due to the nature of the proposed weakly supervised segmentation framework.

\begin{table}
\centering
\caption{Assessment of ICH segmentation performance for Swin-Grad-CAM, Swin-SAM Multi-label, Swin-SAM Binary, Swin-HGI-SAM, U-Net, and Swin-UNETR algorithms, using Dice coefficient and Intersection over Union (IoU). All results are reported as mean±SE. Note the overall metrics are reported based on all cases across the folds.}
\label{tab:seg_results}
\resizebox{\linewidth}{!}{%
\begin{tabular}{|c|c|c|c|c|c|c|} 
\hline
\multirow{2}{*}{\textbf{Fold}} & \multicolumn{6}{c|}{\textbf{Dice Coefficient}} \\ 
\cline{2-7}
 & \textbf{Swin-Grad-CAM} & \begin{tabular}[c]{@{}c@{}}\textbf{Swin-SAM}\\\textbf{Multi-label}\end{tabular} & \begin{tabular}[c]{@{}c@{}}\textbf{Swin-SAM}\\\textbf{Binary}\end{tabular} & \textbf{Swin-HGI-SAM} & \begin{tabular}[c]{@{}c@{}}\textbf{Fully supervised}\\\textbf{U-Net}\end{tabular} & \begin{tabular}[c]{@{}c@{}}\textbf{Fully supervised}\\\textbf{Swin-UNETR}\end{tabular} \\ 
\hline
\textbf{1} & 0.174 ± 0.025 & 0.223 ± 0.029 & 0.337 ± 0.031 & 0.354 ± 0.040 & 0.302 ± 0.045 & 0.281 ± 0.044 \\ 
\hline
\textbf{2} & 0.219 ± 0.024 & 0.319 ± 0.021 & 0.433 ± 0.025 & 0.505 ± 0.026 & 0.336 ± 0.036 & 0.434 ± 0.044 \\ 
\hline
\textbf{3} & 0.189 ± 0.032 & 0.276 ± 0.032 & 0.347 ± 0.033 & 0.414 ± 0.034 & 0.442 ± 0.047 & 0.399 ± 0.045 \\ 
\hline
\textbf{4} & 0.268 ± 0.022 & 0.322 ± 0.025 & 0.400 ± 0.025 & 0.451 ± 0.029 & 0.555 ± 0.036 & 0.571 ± 0.034 \\ 
\hline
\textbf{5} & 0.307 ± 0.026 & 0.338 ± 0.029 & 0.407 ± 0.022 & 0.481 ± 0.030 & 0.491 ± 0.040 & 0.522 ± 0.037 \\ 
\hline
\begin{tabular}[c]{@{}c@{}}\textbf{Overall}\\\textbf{slice-wise}\end{tabular} & \textbf{0.237 ± 0.012} & \textbf{0.299 ± 0.012} & \textbf{0.387 ± 0.012} & \textbf{0.444 ± 0.014} & \textbf{0.438 ± 0.019} & \textbf{0.455 ± 0.019} \\ 
\hline
\multicolumn{7}{|c|}{} \\ 
\hline
\multirow{2}{*}{\textbf{Fold}} & \multicolumn{6}{c|}{\textbf{Intersection over Union}} \\ 
\cline{2-7}
 & \textbf{\textbf{Swin-Grad-CAM}} & \begin{tabular}[c]{@{}c@{}}\textbf{Swin-SAM}\\\textbf{Multi-label}\end{tabular} & \begin{tabular}[c]{@{}c@{}}\textbf{Swin-SAM}\\\textbf{Binary}\end{tabular} & \textbf{\textbf{Swin-HGI-SAM}} & \begin{tabular}[c]{@{}c@{}}\textbf{Fully supervised}\\\textbf{U-Net}\end{tabular} & \begin{tabular}[c]{@{}c@{}}\textbf{Fully supervised}\\\textbf{Swin-UNETR}\end{tabular} \\ 
\hline
\textbf{1} & 0.107 ± 0.017 & 0.143 ± 0.020 & 0.226 ± 0.024 & 0.255 ± 0.031 & 0.228 ± 0.036 & 0.210 ± 0.035 \\ 
\hline
\textbf{2} & 0.136 ± 0.017 & 0.200 ± 0.015 & 0.295 ± 0.020 & 0.360 ± 0.022 & 0.238 ± 0.029 & 0.338 ± 0.038 \\ 
\hline
\textbf{3} & 0.127 ± 0.023 & 0.185 ± 0.024 & 0.240 ± 0.026 & 0.294 ± 0.027 & 0.352 ± 0.041 & 0.308 ± 0.038 \\ 
\hline
\textbf{4} & 0.171 ± 0.015 & 0.214 ± 0.018 & 0.275 ± 0.019 & 0.328 ± 0.024 & 0.453 ± 0.033 & 0.458 ± 0.030 \\ 
\hline
\textbf{5} & 0.199 ± 0.019 & 0.226 ± 0.022 & 0.270 ± 0.018 & 0.347 ± 0.026 & 0.381 ± 0.035 & 0.403 ± 0.033 \\ 
\hline
\begin{tabular}[c]{@{}c@{}}\textbf{Overall}\\\textbf{slice-wise}\end{tabular} & \textbf{0.151 ± 0.008} & \textbf{0.196 ± 0.009} & \textbf{0.263 ± 0.010} & \textbf{0.319 ± 0.012} & \textbf{0.343 ± 0.016} & \textbf{0.355 ± 0.016} \\
\hline
\end{tabular}
}
\end{table}

\begin{table}
\centering
\caption{Assessment of ICH detection performance for Swin-SAM Multi-label, Swin-SAM Binary, Swin-HGI-SAM, U-Net, and Swin-UNETR algorithms, using accuracy, AUC, precision, F1-score, recall, and specificity. Note the overall metrics are reported based on all cases across the folds.}
\label{tab:detection_results}
\resizebox{\linewidth}{!}{%
\begin{tabular}{|c|c|c|c|c|c|c|c|c|c|c|} 
\hline
\multirow{2}{*}{\textbf{Fold}} & \multicolumn{5}{c|}{\textbf{Accuracy}} & \multicolumn{5}{c|}{\textbf{AUC}} \\ 
\cline{2-11}
 & \begin{tabular}[c]{@{}c@{}}\textbf{Swin-SAM}\\\textbf{Multi-label}\end{tabular} & \begin{tabular}[c]{@{}c@{}}\textbf{Swin-SAM}\\\textbf{Binary}\end{tabular} & \textbf{Swin-HGI-SAM} & \textbf{U-Net} & \textbf{Swin-UNETR} & \begin{tabular}[c]{@{}c@{}}\textbf{Swin-SAM}\\\textbf{Multi-label}\end{tabular} & \begin{tabular}[c]{@{}c@{}}\textbf{Swin-SAM}\\\textbf{Binary}\end{tabular} & \textbf{Swin-HGI-SAM} & \textbf{U-Net} & \textbf{Swin-UNETR} \\ 
\hline
\textbf{1} & 0.948 & 0.953 & 0.946 & 0.572 & 0.654 & 0.821 & 0.874 & 0.904 & 0.731 & 0.785 \\ 
\hline
\textbf{2} & 0.958 & 0.964 & 0.958 & 0.751 & 0.816 & 0.851 & 0.891 & 0.902 & 0.830 & 0.845 \\ 
\hline
\textbf{3} & 0.934 & 0.928 & 0.928 & 0.674 & 0.589 & 0.712 & 0.701 & 0.731 & 0.765 & 0.732 \\ 
\hline
\textbf{4} & 0.967 & 0.965 & 0.953 & 0.692 & 0.685 & 0.932 & 0.950 & 0.948 & 0.805 & 0.810 \\ 
\hline
\textbf{5} & 0.959 & 0.954 & 0.938 & 0.515 & 0.660 & 0.935 & 0.939 & 0.937 & 0.721 & 0.788 \\ 
\hline
\begin{tabular}[c]{@{}c@{}}\textbf{Overall}\\\textbf{slice-wise}\end{tabular} & \textbf{0.953} & \textbf{0.953} & \textbf{0.945} & \textbf{0.639} & \textbf{0.679} & \textbf{0.858} & \textbf{0.879} & \textbf{0.891} & \textbf{0.770} & \textbf{0.793} \\ 
\hline
\multicolumn{11}{|c|}{} \\ 
\hline
\multirow{2}{*}{\textbf{Fold}} & \multicolumn{5}{c|}{\textbf{Precision}} & \multicolumn{5}{c|}{\textbf{F1-score}} \\ 
\cline{2-11}
 & \begin{tabular}[c]{@{}c@{}}\textbf{Swin-SAM}\\\textbf{Multi-label}\end{tabular} & \begin{tabular}[c]{@{}c@{}}\textbf{Swin-SAM}\\\textbf{Binary}\end{tabular} & \textbf{Swin-HGI-SAM} & \textbf{U-Net} & \textbf{Swin-UNETR} & \begin{tabular}[c]{@{}c@{}}\textbf{Swin-SAM}\\\textbf{Multi-label}\end{tabular} & \begin{tabular}[c]{@{}c@{}}\textbf{Swin-SAM}\\\textbf{Binary}\end{tabular} & \textbf{Swin-HGI-SAM} & \textbf{U-Net} & \textbf{Swin-UNETR} \\ 
\hline
\textbf{1} & 0.735 & 0.724 & 0.657 & 0.166 & 0.201 & 0.699 & 0.750 & 0.742 & 0.282 & 0.331 \\ 
\hline
\textbf{2} & 0.894 & 0.870 & 0.803 & 0.304 & 0.369 & 0.792 & 0.832 & 0.817 & 0.458 & 0.520 \\ 
\hline
\textbf{3} & 0.862 & 0.800 & 0.737 & 0.226 & 0.191 & 0.575 & 0.545 & 0.583 & 0.359 & 0.315 \\ 
\hline
\textbf{4} & 0.893 & 0.849 & 0.784 & 0.322 & 0.319 & 0.888 & 0.888 & 0.856 & 0.482 & 0.483 \\ 
\hline
\textbf{5} & 0.767 & 0.731 & 0.652 & 0.183 & 0.238 & 0.830 & 0.814 & 0.768 & 0.308 & 0.381 \\ 
\hline
\begin{tabular}[c]{@{}c@{}}\textbf{Overall}\\\textbf{slice-wise}\end{tabular} & \textbf{0.830} & \textbf{0.796} & \textbf{0.725} & \textbf{0.231} & \textbf{0.253} & \textbf{0.780} & \textbf{0.789} & \textbf{0.770} & \textbf{0.370} & \textbf{0.398} \\ 
\hline
\multicolumn{11}{|c|}{} \\ 
\hline
\multirow{2}{*}{\textbf{Fold}} & \multicolumn{5}{c|}{\textbf{Recall (Sensitivity)}} & \multicolumn{5}{c|}{\textbf{Specificity}} \\ 
\cline{2-11}
 & \begin{tabular}[c]{@{}c@{}}\textbf{Swin-SAM}\\\textbf{Multi-label}\end{tabular} & \begin{tabular}[c]{@{}c@{}}\textbf{Swin-SAM}\\\textbf{Binary}\end{tabular} & \textbf{Swin-HGI-SAM} & \textbf{U-Net} & \textbf{Swin-UNETR} & \begin{tabular}[c]{@{}c@{}}\textbf{Swin-SAM}\\\textbf{Multi-label}\end{tabular} & \begin{tabular}[c]{@{}c@{}}\textbf{Swin-SAM}\\\textbf{Binary}\end{tabular} & \textbf{Swin-HGI-SAM} & \textbf{U-Net} & \textbf{Swin-UNETR} \\ 
\hline
\textbf{1} & 0.667 & 0.778 & 0.852 & 0.926 & 0.944 & 0.976 & 0.970 & 0.956 & 0.537 & 0.625 \\ 
\hline
\textbf{2} & 0.712 & 0.797 & 0.831 & 0.932 & 0.881 & 0.989 & 0.985 & 0.974 & 0.728 & 0.808 \\ 
\hline
\textbf{3} & 0.431 & 0.414 & 0.483 & 0.879 & 0.914 & 0.992 & 0.988 & 0.980 & 0.651 & 0.551 \\ 
\hline
\textbf{4} & 0.882 & 0.929 & 0.941 & 0.965 & 0.988 & 0.982 & 0.971 & 0.955 & 0.645 & 0.632 \\ 
\hline
\textbf{5} & 0.903 & 0.919 & 0.935 & 0.984 & 0.952 & 0.966 & 0.958 & 0.938 & 0.457 & 0.624 \\ 
\hline
\begin{tabular}[c]{@{}c@{}}\textbf{Overall}\\\textbf{slice-wise}\end{tabular} & \textbf{0.736} & \textbf{0.783} & \textbf{0.821} & \textbf{0.940} & \textbf{0.940} & \textbf{0.981} & \textbf{0.974} & \textbf{0.960} & \textbf{0.600} & \textbf{0.645} \\
\hline
\end{tabular}
}
\end{table}

\section{Discussion}
In recent years, the urgent need to enhance the transparency of deep learning algorithms has encouraged the development of various techniques to visualize network activation/attention maps in vision tasks. Among them, Grad-CAM \citep{selvaraju2017grad} has gained popularity to reveal the regions of interest in image classification tasks for CNNs, thanks to its simplicity and flexibility. Furthermore, extending its original purpose, it has also been adopted in weakly supervised image segmentation based on categorical and metric learning to generate pixel-level semantic labels  \citep{chen2022c}, including applications for stroke lesion segmentation \citep{Wu2019MICCAI,nemcek2021weakly}. Compared with Grad-CAM and its variants, the more recent attention mechanisms, especially self-attention from transformer models, can identify more discriminative, task-related regions and features while improving the performance of the DL models \citep{LIANG2022206, dosovitskiy2020image}. This was confirmed in this study when comparing the segmentation performance of the proposed weakly supervised ICH segmentation approaches with Grad-CAM and self-attention maps. As for the self-attention mechanism, different learning strategies may influence the positioning and tightness of network attention with respect to the target objects, and thus the downstream segmentation outcomes in the proposed framework. As an ablation study of our previous work \citep{rasoulian2022weakly}, we examined the impact of binary (presence of hemorrhage or not) versus multi-label classifications (ICH subtypes and with/without ICH) on self-attention maps from Swin transformers. Using segmentation accuracy as a metric, we found that binary classification helped the network better focus on the hemorrhage regions while both strategies offer similar performance for ICH detection. In the new segmentation method with HGI-SAM, we followed our earlier insights to build our algorithm. 

Inspired by the popular Grad-CAM technique \citep{selvaraju2017grad}, we incorporated head-wise gradient-weighing for self-attention maps to boost the presentation of the weights relevant to specific class activation for the first time. Compared with other attention mapping techniques \citep{chefer2021transformer,sun2021getam} that relied on the ViT, we were also the first to implement it on the more complex Swin transformer that was intended to improve upon the ViT. The enhanced visualization of the attention maps and ICH segmentation accuracy are evident in Fig.~\ref{attention_comparison} and Fig.~\ref{segmentation_comparison}, respectively. Among the obtained attention maps at different hierarchies, those from the earlier layers contained the relevant attention weights at higher resolutions, and thus were more helpful to delineate the regions of interest (i.e., hemorrhages) through the ICH classification task. In our experiment, the head-wise gradient-infused self-attention map (HGI-SAM) from Layer 4 possesses relatively less discriminative power primarily due to much lower resolution and the adverse cascading effect of transformer architecture \citep{Zhou2021DeepViTTD}. Therefore, we chose to fuse those from the first three layers for our proposed method. In fact, with a few cases, we found that the fused attention map from the first 3 layers offered better segmentation accuracy than using those from all 4 layers. In our original study that relied on self-attention maps alone, fusing all 4 layers was more beneficial. To obtain discrete ICH segmentation from the HGI-SAM, we applied additional post-processing steps. One major procedure that was different from our original article was multiplying the brain mask before ICH mask binarization. When closely inspecting the attention maps from ICH classification, we noticed that skull fractures were also identified in addition to the hemorrhage. This is likely because, for many ICH patients, the condition may result in accidental falls that cause additional injury, such as skull or spine fractures. This phenomenon perfectly showcased the power of attention visualization in explaining the decision-making process in DL models. By constraining the post-processing in the brain region, we intended to exclude the attention weights regarding skull fractures and were able to further improve the segmentation accuracy. Finally, different from our previous approach \citep{rasoulian2022weakly}, where the denoised brain window was multiplied to the attention map, our new method directly performed thresholding on the gradient-weighted map to avoid potential intensity inconsistency within the hemorrhage and multi-center imaging protocols. This also allowed us to directly probe the quality of activation/attention maps with respect to their specificity in focusing on ICH. 

To provide baselines for our weakly supervised segmentation framework, we have trained a U-Net and a Swin-UNETR with full supervision using the PhysioNet data to perform ICH segmentation. By using data sampling to tackle class imbalance in training, our U-Net model has achieved an improved mean Dice score of 0.438 over that of 0.315 reported for the U-Net in the original data paper \citep{hssayeni2020intracranial}. In comparison, our proposed method has achieved similar results to our baseline supervised U-Net and Swin-UNETR ($p>0.05$) with the mean Dice scores slightly lower than the Swin-UNETR and higher than the U-Net, showcasing the feasibility and excellent potential of weakly supervised segmentation with much more accessible categorical labels. In terms of computational cost, U-Net was the most efficient model, taking only around 10ms/sample, likely due to its simple convolutional layer architecture. On the other hand, Swin-UNETR took around 15ms/sample, Swin-SAM models took around 30ms/sample, and Swin-HGI-SAM and Swin-Grad-CAM took approximately 60ms and 90ms per sample, respectively. 
The longer inference time is because the latter two required backward operations for gradient computation, which is a key step for the proposed framework. However, all these models are still relatively fast and suitable for clinical setups, offering practical benefits. 

While there is still room for improvement in our future work, ICH segmentation from clinical scans remains a challenging task at the moment. In our proposed framework, extracting meaningful pixel-wise attention maps is crucial. 
We admit that the exploration of self-attention in this study may be data-, application- and model-specific while the baseline supervised models have been tested in various applications. By using categorical learning to obtain attention and saliency maps for segmentation, depending on the data and application, it is possible that the local regions that the network focuses on for image classification may not fully overlap with the segmentation ground truths. In our application, the derived self-attention maps focused on both ICH lesions and skull fractures in some cases, and we used skull stripping to tackle this. In the future, we will continue to investigate the characteristics of self-attention in different learning strategies, extended applications, and other Transformer models. These would be greatly beneficial to improve weakly supervised medical image segmentation based on categorical labels. Incorporating inter-slice or 3D spatial information may be beneficial to the designated tasks, especially for 2D slices that contain a few pixels of ICH, but the high variability of CT slice thickness in the public datasets has posed challenges in the 3D approach. Recent developments in resolution-agnostic brain image segmentation \citep{BILLOT2023} and image super-resolution \citep{Sui2021} through generative DL models have allowed high-quality interpretation of clinical scans with diverse imaging protocols (e.g., different image resolutions). We will seek to adapt these frameworks for CT images in the task of ICH detection and segmentation in future work. 

\section{Conclusion}
To mitigate the requirement of expensive training data for intracranial hemorrhage segmentation, we have proposed a weakly supervised framework by using a novel hierarchical combination of head-wise gradient-infused self-attention maps from a Swin transformer through categorical learning. By using two public CT databases, we further demonstrated the benefits of head-wise gradient-weighing of derived attention maps to further boost ICH segmentation performance for the first time. In the future, we will further explore the proposed HGI-SAM technique and the application of the proposed weakly supervised segmentation framework in extended applications and other Transformer models.

%%%%%%%%%%%%%%%%%%%%%%%%%%%%%%%%%%%%%%%%%%%%%%%%%%%%%%%%%%%%%%%%%%%%%%%
% Mandatory Sections. Please complete, especially for final publication
%%%%%%%%%%%%%%%%%%%%%%%%%%%%%%%%%%%%%%%%%%%%%%%%%%%%%%%%%%%%%%%%%%%%%%%

% Acknowledgements.
% Please include any funding, intellectual contributions not included in the authorship, and any other acknowledgements.
\acks{This work was supported by a Fonds de recherche du Quebec - Nature et technologies (FRQNT) Team Research Project Grant (2022-PR296459).}

% Ethical Standards.
% Please edit with the appropriate ethics considerations for your work. Include any pertinent IRB information, etc.
%
% Please note that the submission requirements included:
% The work presented must follow appropriate ethical standards in conducting research and writing the manuscript, following all applicable laws and regulations regarding treatment of animals or human subjects.
\ethics{The work follows appropriate ethical standards in conducting research and writing the manuscript, following all applicable laws and regulations regarding treatment of animals or human subjects.}

% Conflict of Interest
% Declaration of possible conflicts of interest: Authors must disclose any financial, organisational, commercial or personal conflicts of interest that might bias their work.
% If no conflicts, please say "We declare we don't have conflicts of interest."
\coi{The authors declare no conflicts of interest for this work.}

\bibliography{sample}

% needed in general before appendices.
% \newpage

% Appendix is optional
\clearpage
\appendix

\end{document}